\documentclass[lettersize,journal]{IEEEtran}
\usepackage{amsmath,amsfonts,amsthm,amsbsy}
\usepackage{algorithmic}
\usepackage{algorithm}
\usepackage{array}
\usepackage[caption=false,font=normalsize,labelfont=sf,textfont=sf]{subfig}
\usepackage{textcomp}
\usepackage{stfloats}
\usepackage{url}
\usepackage{verbatim}
\usepackage{graphicx}
\usepackage{cite}
\usepackage{hyperref}
\usepackage{titlesec}
\usepackage{bm}
\usepackage{mathtools}
\usepackage{xcolor}
\newcommand{\EdgeSetP}{\vec{\bm{\mathcal{E}}}_p}
\newcommand{\EdgeSetB}{\vec{\bm{\mathcal{E}}}_b}
\newcommand{\EdgeSetS}{\vec{\bm{\mathcal{E}}}_s}
\newcommand{\EdgeSet}{\vec{\bm{\mathcal{E}}}}
\newcommand{\VertSetP}{\bm{\mathcal{V}}_p}
\newcommand{\VertSetM}{\bm{\mathcal{V}}_m}

\newcommand{\VertSetS}{\bm{\mathcal{V}}_s}
\newcommand{\VertSet}{\bm{\mathcal{V}}}

\newcommand{\GraphB}{\vec{\bm{\mathcal{G}}}_b}
\newcommand{\GraphS}{\vec{\bm{\mathcal{G}}}_s}
\newcommand{\Graph}{\vec{\bm{\mathcal{G}}}}
\newcommand{\card}[1]{\vert#1\vert}
\newcommand{\half}{\frac{1}{2}}
\newcommand{\E}[1]{\times10^{#1}}

\titlespacing*{\subsection}{0pt}{0.1\baselineskip}{0.1\baselineskip}
\titlespacing*{\section}{0pt}{0.2\baselineskip}{0.1\baselineskip}
\begin{document}

\title{An Efficient Global Optimality Certificate for Landmark-Based SLAM}

\author{Connor Holmes and Timothy D. Barfoot\vspace*{-0.45in}
\thanks{This work was generously supported by the National Sciences and Engineering Research Council of Canada (NSERC).}
\thanks{Connor Holmes and Timothy D. Barfoot are with the University of Toronto Robotics Institute, University of Toronto, Toronto, Ontario, Canada, \texttt{connor.holmes@mail.utoronto.ca}, \texttt{tim.barfoot@utoronto.ca}.}}%
\pagenumbering{gobble}


\maketitle

\begin{abstract}

Modern state estimation is often formulated as an optimization problem and solved using efficient local search methods. These methods at best guarantee convergence to local minima, but, in some cases, global optimality can also be certified. Although such global optimality certificates have been well established for 3D \textit{pose-graph optimization}, the details have yet to be worked out for the 3D landmark-based SLAM problem, in which estimated states include both robot poses and map landmarks. 
In this paper, we address this gap by using a graph-theoretic approach to cast the subproblems of landmark-based SLAM into a form that yields a sufficient condition for global optimality. Efficient methods of computing the optimality certificates for these subproblems exist, but first require the construction of a large data matrix. We show that this matrix can be constructed with complexity that remains linear in the number of landmarks and does not exceed the state-of-the-art computational complexity of one local solver iteration.
We demonstrate the efficacy of the certificate on simulated and real-world landmark-based SLAM problems.
We also integrate our method into the state-of-the-art SE-Sync pipeline to efficiently solve landmark-based SLAM problems to global optimality.
Finally, we study the robustness of the global optimality certificate to measurement noise, taking into consideration the effect of the underlying measurement graph.

\end{abstract}


\section{Introduction}
\IEEEPARstart{S}{tate} estimation for traditional robotic systems has become a well-established field. State-of-the-art, workhorse algorithms for state estimation problems, such as Simultaneous Localization and Mapping (SLAM), are now capable of estimating hundreds of thousands of states on a single processor in real time \cite{rosen2021advances}. Such algorithms have reached a high level of maturity in terms of both breadth and depth of understanding. See \cite{durrantwhyte06a} and \cite{durrantwhyte06b} for a comprehensive review of SLAM. 

In recent years, the focus of state-estimation research has shifted to the improvement of robustness of these modern algorithms. Indeed, the majority of such algorithms depend on fast, local optimization of non-convex problems and are therefore susceptible to convergence to local (but not global) stationary points. To address this issue, efforts to understand more deeply the convergence properties of SLAM have emerged. The current state of these efforts, which are centered around the so-called \textit{certifiably correct SLAM} methods, are well surveyed in \cite{cadena2016pastSurvey} and \cite{rosen2021advances}.



\begin{figure}[!t]
	\centering
	\includegraphics[width=\columnwidth]{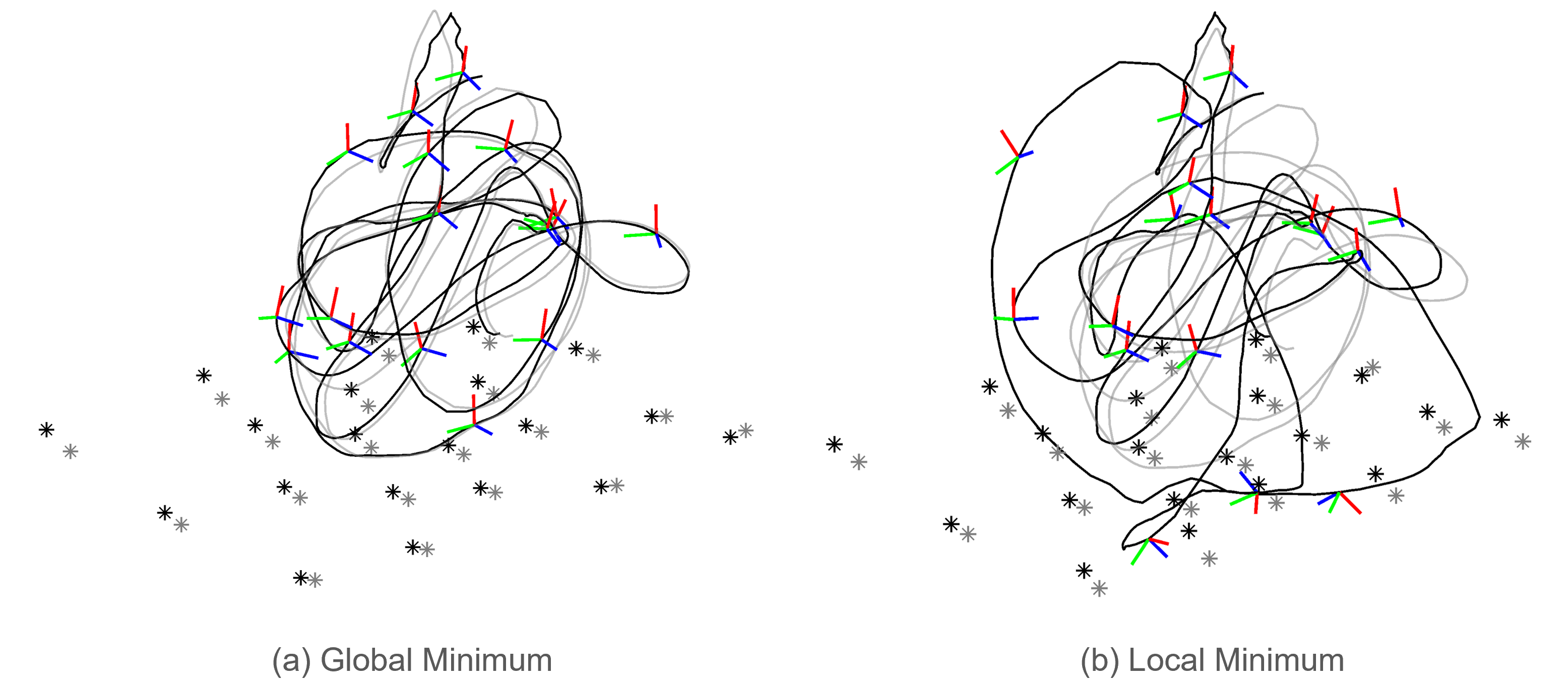}
	\vspace*{-0.3in}
	\caption{Two \emph{converged} solutions to the landmark-based SLAM problem on the ``Starry Night'' dataset, based on different initial guesses. The global optimality test successfully identifies the global minimum and the local minimum solution. Landmarks are represented as black stars and the pose trajectory is represented by the black line (subset of poses shown as RGB frames). Light gray trajectory and stars represent the ground truth poses and landmarks, respectively.}
	\vspace*{-10pt}
	\label{fig:starry}
\end{figure}

One of the earliest adoptions of certifiably correct methods was the certification for a subproblem of SLAM, pose-graph optimization (PGO), in \cite{carlone2016planar}\cite{carlone2015lagrangian}. In these papers, the problem was cast as a quadratically constrained quadratic program (QCQP) whose Lagrangian dual problem, a semidefinite program (SDP), was often tight in practice. \cite{rosen2019se} extended these results by studying the SDP relaxation of the PGO problem and proved that globally optimal solutions could be found in problems with sufficiently low levels of measurement noise. Improvements to these relaxation methods (including a distributed approach) have been further explored in \cite{briales2017cartan, tian2021distributed}. 

Another key robotics and vision problem is \textit{multiple point cloud registration} (MPCR) for which exact and stable global solutions were studied in \cite{chaudhury_global_2015}, again using an SDP relaxation. 
More recently, \cite{iglesias_global_2020} used Lagrangian duality to further investigate the conditions under which solutions to this problem could be globally certified. 

Lagrangian dual methods have been studied at length in terms of the problem of rotational averaging (RA) \cite{wilson2016rotations, eriksson2018rotation, eriksson2019rotation, dellaert2020shonan}. Similar methods have also been applied to certification of \textit{robust} state-estimation \cite{carlone2018convex,yang2020teaser}, sensor calibration \cite{wise2020certifiably,giamou2019certifiably} and image segmentation \cite{hu2019accelerated}, among other QCQP perception problems. Critically, \cite{cifuentes2021local} shows that, under certain conditions, QCQP problems have zero duality gap when unperturbed (no noise) and continue to enjoy zero duality gap as long as the perturbation parameter is within a bound (i.e., the underlying problem has sufficiently low noise). For state-estimation problems, this bound is typically larger than practical noise levels.

Of interest here is the combination of MPCA and PGO problems, which are together known as \textit{landmark-based SLAM}. Certifiably optimal solutions have already been studied for the 2D version of this problem using a complex number representation \cite{fan_cpl-slam_2020}. However, a global certificate for the 3D \textit{landmark-based SLAM} problem has not yet been addressed in the literature and is the focus of this paper. While the expert reader may recognize that the SE-Sync algorithm \cite{rosen2019se} can, in theory, already be applied to this problem by treating landmarks as degenerate poses, this can lead to a considerable increase in the size of the problem that must be solved. Instead, we explicitly incorporate landmarks into the problem statement and, in doing so, can \emph{efficiently} integrate landmarks into the existing SE-Sync pipeline. This also provides a convenient formulation for the investigation of the effect of landmark measurements on optimality certificate complexity and on robustness of the certificate to noise. 

The first contribution of this paper is to use a graph-theoretic approach (in a manner similar to \cite{chaudhury_global_2015} and \cite{rosen2019se}) to cast the landmark-based SLAM problem (and its subproblems) into a specific QCQP form (Sections \ref{Sec:Problem} and \ref{Sec:SLAMForm}). In Section \ref{Sec:OptCert}, we review sufficient conditions for global optimality can be computed for this QCQP. As the second technical contribution of this paper, we show that our global optimality test can scale linearly with landmark measurements, ensuring efficient computation even when the number of landmarks is very large. In Section \ref{Sec:NumRes} we demonstrate the efficacy of the optimality test on simulated and real-world datasets. In Section \ref{Sec:SeSyncCompl}, we show the runtime results of the integration our method into SE-Sync and demonstrate numerical proof of our complexity claims. Finally, in Section \ref{Sec:NoiseDuality} we investigate certificate robustness to measurement noise, taking into consideration the effect of the \emph{structure} of the measurements. 

\section{Notation} \label{Sec:Notn}
We denote matrices with bold-faced, capitalized letters, $ \bm{A} $, column vectors with bold-faced, lower-case letters, $ \bm{a} $, and scalar quantities normal-faced font, $ a $.
Let $ \mathbb{S}^n $ denote the space of $ n $-dimensional symmetric matrices and $ \mathbb{S}_+^n $ denote the space of $ n $-dimensional symmetric positive-semidefinite matrices.
Let $ \|\cdot\|_F $ denote the Frobenius norm.
Let $ \mbox{diag}(\bm{A}_1,\dots,\bm{A}_N) $ denote the block-diagonal matrix with blocks corresponding to matrices, $ \bm{A}_1,\dots,\bm{A}_N $. Note that this includes the case where the $ \bm{A}_i $ are scalar.
Let $ \bm{I} $ denote the identity matrix with dimension three unless otherwise specified.
Let $ \bm{0} $ denote the matrix with all-zero entries, whose dimension will be evident from the context.
Let the subscript $ o $ denote the world frame.
Let $ \bm{r}_i^{ji} $ denote a vector from frame $ i $ to frame $ j $ expressed in frame $ i $ and $ \bm{R}_{ij} $ denote a rotation matrix that maps vectors expressed in frame $ j $ to equivalent vectors in frame $ i $.
Let $ \bar{\bm{B}} = \bm{B} \otimes \bm{I}$, where $ \bm{B} $ is any matrix, $ \bm{I} $ is the $ 3\times3 $ identity matrix, and $ \otimes $ denotes the Kronecker product.
Let $ \vert S \vert $ denote the cardinality of the set $ S $.
Let $ \bm{A}^+ $ denote the Moore-Penrose pseudoinverse of a given matrix $ \bm{A} $.
Let $ \mbox{vec}(\bm{A}) $ denote the vectorization (reshape) of a given matrix $ \bm{A} $.

Suppose we have a directed graph $ \Graph = \{\VertSet, \EdgeSet\}$. We define the \textit{incidence matrix}, $ \bm{B} = \left[\bm{B}_{ie}\right] $, of $ \Graph $ element-wise for each edge $ e \in \EdgeSet $,
\begin{equation*}
	\bm{B}_{ie} = \left\{
	\begin{array}{ll}
		1 \quad & \mbox{if } e = (*,i) \\
		-1 \quad &  \mbox{if } e = (i,*) \\
		0 \quad &\mbox{otherwise}
	\end{array} \right. \quad \forall i \in \VertSet\quad. 
\end{equation*}
We also define the \textit{restricted incidence matrix}, $ \bm{B}^p =\left[\bm{B}^p_{ie}\right]$, as the incidence matrix of outgoing edges from the subgraph with vertex set restricted to $ \VertSetP \subset \VertSet $:
\begin{equation*}
	\bm{B}^p_{ie} = \left\{
	\begin{array}{ll}
		1 \quad & \mbox{if }  e = (i,*) \\
		0 & \mbox{otherwise}
	\end{array} \right. \quad \forall i \in \VertSetP \subset \VertSet \quad.
\end{equation*}

\section{Problem Setup} \label{Sec:Problem}

In this paper, we consider a landmark-based SLAM problem in three dimensions, which consists of estimating $ N_p $ robot poses (positions, $ \bm{t}_o^{io} \in\mathbb{R}^3 $, and orientations, $ \bm{R}_{oi}\in\mbox{SO}(3) $) as well as $ N_m $ landmark positions, $ \bm{m}_o^{jo} \in\mathbb{R}^3 $. 
We consider the simplified case of the 3D SLAM problem, in which landmark measurements are 3D points in the observing pose's frame, rather than outputs of a nonlinear measurement model. We also do not consider outlier measurements.
We associate a given SLAM problem with a directed \textit{SLAM graph}, $ \GraphS = \{\VertSetS,\EdgeSetS\}$, having vertex set $ \VertSetS = \VertSetP \cup \VertSetM $, where $ \VertSetP $ represents poses ($ \card{\VertSetP}=N_p $) and $ \VertSetM $ represents map landmarks ($ \card{\VertSetM}=N_m $).

Each edge of $ \GraphS $ is associated with a \textit{measurement} or \textit{piece of information} that we have in the SLAM problem. In particular, we define the edge set, $ \EdgeSetS = \EdgeSetB \cup \EdgeSetP $, where each edge, $ e = (i,j) \in \EdgeSetB $, is associated with a pose-to-landmark measurement, $ \tilde{\bm{y}}_i^{ji}\in \mathbb{R}^3 $, from pose $ i $ to landmark $ j $.
Similarly, each edge,  $e = (i,k) \in \EdgeSetP $, is associated with measurements, $ \tilde{\bm{t}}_i^{ki} \in \mathbb{R}^3$ and $ \tilde{\bm{R}}_{ik} \in SO(3)$, from pose $ i $ to pose $ k $.

We define the landmark-based SLAM optimization problem as follows:
\begin{equation}\label{opt:SLAM}
	\begin{array}{rll}
		\mbox{min} &  J_r + J_b + J_t&\\
		\mbox{w.r.t.} & \bm{R}_{oi}, \bm{t}_o^{io},\bm{m}_o^{jo} \quad& (\forall i \in \VertSetP,\forall j \in \VertSetM),\\
		\mbox{s.t.} &  \bm{R}_{oi}\bm{R}_{oi}^T = \bm{I} \quad& (\forall i \in \VertSetP),
	\end{array}
\end{equation}
where each of the three cost functions represents a SLAM subproblem,
\begin{subequations}
\begin{gather}
	J_r = \sum_{(i,k)\in \EdgeSetP} w^r_{ik} \left\| \bm{R}_{oi} \widetilde{\bm{R}}_{ik} - \bm{R}_{ok} \right\|_F^2, \label{eqn:RAcost}\\
	J_b = \sum_{(i,j)\in\EdgeSetB} w^b_{ij} \left\| \bm{R}_{oi} \tilde{\bm{y}}_i^{ji} - \left( \bm{m}_o^{jo} - \bm{t}_o^{io} \right)\right\|_2^2, \label{eqn:BAcost}\\
	J_t = \sum_{(i,k)\in \EdgeSetP} w^t_{ik} \left\| \bm{R}_{oi} \widetilde{\bm{t}}_i^{ki} - \left( \bm{t}_o^{ko} - \bm{t}_o^{io} \right)\right\|_2^2, \label{eqn:TScost}
\end{gather}
\end{subequations}
and $  w^b_{ij} $, $  w^r_{ik} $ and $  w^t_{ik} $ are scalar weights. 
These costs are typically associated with Gaussian-distributed noise for the translation errors and the Langevin-distributed noise for the rotational errors \cite{rosen2019se}. 
In the global optimization literature for robotics, these subproblems have been (separately) studied extensively and are referred to as \emph{rotational averaging} (RA) \eqref{eqn:RAcost}, \emph{multiple point cloud registration} (MPCR) \eqref{eqn:BAcost}, and \emph{pose-graph optimization} (PGO) (\ref{eqn:RAcost} and \ref{eqn:TScost}). One of the contributions of this paper is the demonstration of a general expression that encompasses all of these problems.

Note that in Problem \eqref{opt:SLAM}, we have relaxed the typical $ SO(3) $ membership constraints on pose rotations to constraints with membership in $ O(3) $:
\begin{equation*}
	\bm{R}_{oi} \in O(3) \iff \bm{R}_{oi}^T\bm{R}_{oi} = \bm{R}_{oi}\bm{R}_{oi}^T = \bm{I}, ~ (\forall i = 1,\dots,N_p).
\end{equation*} 
This relaxation is common in the robotics and vision literature since a solution in $ \mbox{SO}(3) $ can easily be recovered from a solution in $ \mbox{O}(3) $.

We seek the global minimizer of Problem \eqref{opt:SLAM}. However, this problem is a high-dimensional, non-convex optimization that is typically solved efficiently using local optimization methods (such as Gauss-Newton), which can converge to local minima.

\section{SLAM Formulation} \label{Sec:SLAMForm}

In the following subsections, we reformulate the costs of \eqref{opt:SLAM} into a common quadratic form. We then combine the subproblems into a generic QCQP that is known to admit conditions for global optimality. 

\subsection{Relative Rotation Cost Functions}
We begin by rewriting the RA cost function given in \eqref{eqn:RAcost} as
\begin{equation*}
	J_r =2\sum_{(i,k)\in \EdgeSetP} w^r_{ik} \mbox{tr}\left( \bm{I} - \tilde{\bm{R}}_{ik}^T\bm{R}_{oi}^T\bm{R}_{ok} \right) = \mbox{tr} \left(\bm{Q}_r \bm{R}^T \bm{R}\right),
\end{equation*}
where we have used the fact that $ \tilde{\bm{R}}_{ik} $, $ \bm{R}_{oi} $, and $ \bm{R}_{ok} $ are orthogonal matrices. The rotation variables have been collected into a convenient matrix 
$	\bm{R} = \begin{bmatrix}
		\bm{R}_{o1} & \cdots & \bm{R}_{oN_p}
	\end{bmatrix},
$ 
and we have defined the \textit{rotational averaging data matrix}, $ \bm{Q}_r =\left[\bm{Q}_{r,ik}\right] $, as follows:
\begin{equation}\label{eqn:Qr}
	\bm{Q}_{r,ik} = \left\{\begin{array}{ll}
		\sum\limits_{(i,l)\in \EdgeSetP} 2~w^r_{il}~\bm{I}, \quad&\mbox{if } i = k \\
		-\tilde{\bm{R}}_{ik}w^r_{ik}, & \mbox{if } (i,k) \in \EdgeSetP \\
		-\tilde{\bm{R}}_{ik}^Tw^r_{ik}, & \mbox{if } (k,i) \in \EdgeSetP
	\end{array}\right. .
\end{equation}
Note that this definition is consistent with the data matrix definition given in \cite{rosen2019se}. 

\subsection{Translation Cost Functions}\label{Sec:Trans}

In this section, we address the costs associated with the translation subproblems, \eqref{eqn:BAcost} and \eqref{eqn:TScost}. Note that the form of the summands of these subproblems is very similar and can be succinctly combined into a single cost.
To this end, we select a fixed ordering for the edge set, $ \EdgeSetS $, such that for any $ e_b \in \EdgeSetB $ and $ e_p \in \EdgeSetP $, we have $ e_b < e_p $. Next, we collect the optimization variables and measurements into matrices as follows:
\begin{subequations}
	\begin{equation}
		\bm{M} = 
		\begin{bmatrix}
			\bm{t}_o^{1o} \cdots \bm{t}_o^{N_p o} ~ \bm{m}_o^{1o}\cdots \bm{m}_o^{N_m o}
		\end{bmatrix},
	\end{equation}
	\begin{equation}
		\bm{Y}_s = \mbox{diag}(\tilde{\bm{y}}_1,\dots,\tilde{\bm{y}}_{\card{\EdgeSetB}},\tilde{\bm{t}}_1,\dots,\tilde{\bm{t}}_{\card{\EdgeSetP}}),
	\end{equation}
\end{subequations}
where $ \tilde{\bm{y}}_e = \tilde{\bm{y}}_i^{ji}$ for $ e = (i,j) \in \EdgeSetB $ and $ \tilde{\bm{t}}_i = \widetilde{\bm{t}}_i^{ki} $ for  $ e = (i,k) \in \EdgeSetP $. Using these matrices, an element $ J_{bt,e} $ of the sum $ J_{bt}=J_b + J_t $ can be rewritten as follows:
\begin{equation*}
	\begin{split}
		J_{bt,e} &= w_e \left\| \bm{R}(\bm{B}^p_{e} \otimes \bm{I}) \bm{Y}_s - \bm{M} \bm{B}_{e} \right\|_2^2 \\
		&=  \left\| \bm{R}( \bm{B}^p_{e} w^{\half}_e \otimes \bm{I}) \bm{Y}_s - \bm{M} (\bm{B}_{e}w^{\half}_e) \right\|_2^2 
		 = \left\| \bm{E}_{e} \right\|_2^2,
	\end{split}
\end{equation*}
where $ w_e $ represents either $ w^b_{ij} $ or $ w^t_{ik} $ depending on edge $ e \in \EdgeSetS $, and $ \bm{B}_{e} $ and $ \bm{B}^p_{e} $ are the columns of the incidence matrix and restricted incidence matrix of $ \GraphS $, respectively, corresponding to edge $ e \in \EdgeSetS $. We can stack these error vectors column-wise into an error matrix, $ \bm{E}=\left[\bm{E}_{1} \cdots \bm{E}_{\card{\EdgeSetS}}\right] $, and express $ J_{bt} $ using the Frobenius norm,
\begin{equation}\label{eqn:costCombTrans}
	J_{bt} = \left\| \bm{E}\right\|_F^2 = \left\| \bm{R}\bm{\bar{V}}^p_{s} \bm{Y}_s - \bm{M} \bm{V}_s \right\|_F^2,
\end{equation}
where $ \bm{V}_{s} = \bm{B}_s \bm{W}_{s}^{\half}$ and $ \bm{V}^p_{s} = \bm{B}^p_s \bm{W}_s^{\half}$ are the \textit{weighted} incidence and restricted incidence matrices, with weight matrix $ \bm{W}_s = \mbox{diag}(w^b_1,\dots, w^b_{\card{\EdgeSetB}}, w^t_1,\dots, w^t_{\card{\EdgeSetP}})$ and the overbar is consistent with Section \ref{Sec:Notn}. 

According with the edge ordering given above, the incidence matrix for $ \GraphS $ is given by
\begin{equation}\label{eqn:Bmat}
	\bm{B}_s = \begin{bmatrix}
		\bm{B}_{bp} & \bm{B}_{tp} \\
		\bm{B}_{bm} & \bm{0}
	\end{bmatrix},
\end{equation}
where $ \bm{B}_{bp} $, $ \bm{B}_{bm} $, and $ \bm{B}_{tp} $ represent row and column partitions of $ \bm{B}_{s} $ with respect to the vertex sets, $ \VertSetP $ and $ \VertSetM $, and edge sets, $ \EdgeSetB$ and $ \EdgeSetP $, respectively. 

We note that restricting the cost, $ J_{bt} $, to either $ J_t $ or $ J_b $ is equivalent to removing the appropriate columns of $ \bm{B}_s $ along with the corresponding columns and rows of $ \bm{Y}_s $ and $ \bm{W}_s $. This restriction does not affect the development in subsequent sections.

\subsection{Projection of Translation Variables}

The combined cost function defined by \eqref{eqn:costCombTrans} is a function of two matrix variables, $ \bm{R} $ and $ \bm{M} $. Although $ \bm{R} $ is constrained by the $ SO(3) $ constraints in Problem \eqref{opt:SLAM}, $ \bm{M} $ is unconstrained. Therefore, we can solve the \textit{unconstrained} optimization in $ \bm{M} $ to obtain a new optimization problem that is exclusively in terms of $ \bm{R} $, similar to the so-called \textit{variable projection method} \cite{golub2003separable,khosoussi2015exploiting}. Note that this method is also used in \cite{rosen2019se} and \cite{wise2020certifiably}, among others. Taking the partial derivative of the cost $ J_{bt} $ with respect to $ \bm{M} $ (see \cite{magnus2019matrix} for details on matrix calculus), we arrive at the following first-order condition for optimal $ \bm{M}^* $:
\begin{equation*}
	\bm{0} = \left.\frac{\partial J_{bt}}{\partial\bm{M}}\right|_{\bm{M}^*} = \bm{M}^*\bm{L}_s - \bm{R} \bar{\bm{V}}^p_s \bm{Y}_s \bm{V}^T_s,
\end{equation*}
where $ \bm{L}_s = \bm{V}_s\bm{V}^T_s = \bm{B}_s\bm{W}_s\bm{B}^T_s $ is the \textit{weighted Laplacian matrix} of the graph $ \GraphS $.  It is well known that the minimum eigenvalue of the Laplacian matrix of a weighted graph is equal to zero, with eigenvector equal to the vector of ones, $ \begin{bmatrix} 1 \cdots 1 \end{bmatrix}^T $. Therefore, the optimal solution for $ \bm{M} $ (for any given matrix $ \bm{R} $) is given by
\begin{equation}\label{eqn:transSoln}
	\bm{M}^* = \bm{R} \bar{\bm{V}}^p_s \bm{Y}_s \bm{V}^T_s \bm{L}^+_s + \bm{M}_0,~~ \bm{M}_0 = \begin{bmatrix} 1 \cdots 1 \end{bmatrix} \otimes \bm{m}_0,
\end{equation}
where $ \bm{m}_0 \in \mathbb{R}^3$ is a free variable that represents a translation added uniformly to all pose translations and landmark locations. This variable represents the \textit{translational gauge freedom} inherent within the problem. Without loss of generality, we assume that $ \bm{m}_0 = \bm{0} $ in the sequel. 

Substituting this solution into \eqref{eqn:costCombTrans}, the cost can now be written as
\begin{equation*}
	J_{bt} = \left\| \bm{R}\bar{\bm{V}}^p_{b} \bm{Y}_s (\bm{I} - \bm{V}^T_s \bm{L}^+_s \bm{V}_s) \right\|_F^2 = \left\| \bm{R}\bar{\bm{V}}^p_{b} \bm{Y}_s \bm{A}_s \right\|_F^2,
\end{equation*}
where we have used the fact that $ \bm{X}^+ = \bm{X}^T(\bm{X}\bm{X}^T)^+ $. The matrix $\bm{A}_s = (\bm{I} - \bm{V}^+_s \bm{V}_s) $ is the projection matrix into the nullspace of the weighted incidence matrix, $ \bm{V}_s $. In the parlance of algebraic graph theory, this space is connected to the \textit{cycle space} of $ \GraphS $, whose basis generates all possible closed paths in a graph\cite{godsil_cycle_2016}.
We now express the Frobenius norm in terms of the trace,
\begin{equation*}
		J_{bt} = \mbox{tr}\left( \right(\bm{R}\bar{\bm{V}}^p_{s} \bm{Y}_s \bm{A}_s\left)^T \right(\bm{R}\bar{\bm{V}}^p_{s} \bm{Y}_s \bm{A}_s\left) \right)
		= \mbox{tr}\left( \bm{Q}_{bt}\bm{R}^T\bm{R} \right),
\end{equation*}
where we have defined the positive-semidefinite \textit{data matrix}, $ \bm{Q}_{bt} = \bar{\bm{V}}^p_{s} \bm{Y}_s \bm{A}_s\bm{Y}_s^T\bar{\bm{V}}^{pT}_{s} $, for the translation subproblems using the fact that $ \bm{A}_s $ is idempotent and symmetric.

As mentioned above, in the case that the cost is restricted to either $ J_b $ or  $ J_t $, the preceding development still holds and respective data matrices, $ \bm{Q}_b $ and $ \bm{Q}_t $, are given by:
\begin{equation}
	\bm{Q}_b = \bar{\bm{V}}^p_{b} \bm{Y}_b \bm{A}_b\bm{Y}_b^T\bar{\bm{V}}^{pT}_{b}, \quad
	\bm{Q}_t = \bar{\bm{V}}^p_{t} \bm{Y}_t \bm{A}_t\bm{Y}_t^T\bar{\bm{V}}^{pT}_{t},
\end{equation}
where $\bm{A}_b = (\bm{I} - \bm{V}^+_b \bm{V}_b) $, $\bm{A}_t = (\bm{I} - \bm{V}^+_t \bm{V}_t) $, and subscripts correspond to restrictions to partitions defined for \eqref{eqn:Bmat} in Section \ref{Sec:Trans}. 

\subsection{General Formulation}
In the preceding sections, we have  demonstrated that all of the subproblems outlined in this paper can be expressed via the same QCQP form, which, as we will show in Section \ref{Sec:OptCert}, leads directly to a sufficient condition for global optimality of a locally optimal solution. Therefore, the landmark-based SLAM problem \eqref{opt:SLAM} (and any of its subproblems) can be expressed as follows:
\begin{equation}\label{opt:GenSDP}
	\begin{array}{rl}
		\mbox{min} & p(\bm{R})=\mbox{tr} \left(\bm{Q} \bm{R}^T \bm{R}\right)\\
		\mbox{w.r.t.} & \bm{R} =
		\begin{bmatrix}
			\bm{R}_{o1} & \cdots & \bm{R}_{oN_p}
		\end{bmatrix}\\
		\mbox{s.t.} & \bm{R}_{oi}\bm{R}_{oi}^T = \bm{I}, \quad (\forall i = 1,\dots,N_p)
	\end{array} ,
\end{equation}
where the data matrix $ \bm{Q} $ depends on the problem that is being solved:
\begin{equation*}
	\bm{Q} = \left\{\begin{array}{lcl}
		\bm{Q}_r & : & \mbox{Rotational Averaging} \\
		\bm{Q}_r + \bm{Q}_t & : & \mbox{Pose-Graph Optimization} \\
		\bm{Q}_b & : &  \mbox{Multiple Point Cloud Alignment} \\
		\bm{Q}_{bt} & : & \mbox{MPCA with Relative Translations} \\
		\bm{Q}_{b} + \bm{Q}_r & : & \mbox{MPCA with Relative Rotations} \\
		\bm{Q}_{bt} + \bm{Q}_r & : & \mbox{Landmark-Based SLAM}
	\end{array}
	\right. .
\end{equation*}
As noted in \cite{briales2017cartan}, all of the available relative rotation information of each problem can be condensed into an underlying matrix-weighted graph whose \textit{Connection Laplacian} is exactly the matrix $ \bm{Q} $.

\section{Global Optimality Certificates}\label{Sec:OptCert}

Global optimality of solutions to problems in the form of Problem \eqref{opt:GenSDP} have been studied extensively in both the robotics \cite{aholt2012qcqp,rosen2019se, eriksson2018rotation,iglesias_global_2020} and the optimization \cite{cifuentes2021local,lasserre_global_2001,parrilo_semidefinite_2003,shor1987quadratic} literature. A general approach to obtaining an optimality certificate involves first deriving the Lagrangian dual problem and then finding conditions under which \textit{strong duality} holds at a given solution, which then guarantees global optimality (see \cite{boyd2004convex} for details of these concepts). 
\subsection{Lagrangian Dual and Certificates}\label{Sec:LagrDualCert}
In this section, we introduce the Lagrangian dual and its connection with optimality certificates. This connection has been well studied in the aformentioned literature. Therefore, in the interest of brevity, we have moved the derivation of these problems to Appendix \ref{App:LagrDual} and present only the key ideas. 

The convex, dual SDP associated with Problem \eqref{opt:SLAM} is given by
\begin{equation}\label{opt:GenDual}
	\max\limits_{\bm{\Lambda}} d(\bm{\Lambda}) =\mbox{tr}(\bm{\Lambda}), \quad \mbox{s.t.}~ \bm{H} \coloneqq(\bm{Q}-\bm{\Lambda}) \succeq \bm{0},
\end{equation}
where the Lagrange multiplier matrix, $ \bm{\Lambda} $, has a specific, block-diagonal structure  $ \bm{\Lambda} = \mbox{diag}(\bm{\Lambda}_1,\dots,\bm{\Lambda}_{N_p})$. Note that strong duality for the original QCQP \eqref{opt:GenSDP} holds whenever the corank of the matrix $ \bm{H}_d \coloneqq \bm{Q} - \bm{\Lambda}^*_d $ is 3, where $ \bm{\Lambda}^*_d $ represents the optimal dual variable. We will use this property in Section \ref{Sec:NoiseDuality} when analyzing the effect of noise on strong duality for landmark-based SLAM.

Consider a candidate solution, $ \bm{R}^* $, to Problem \eqref{opt:GenSDP}, which has been computed with a fast, local solver and is potentially a global minimum. One potential test for global optimality is to use off-the-shelf SDP solvers \cite{cvx,mosek} to find the optimal dual variables of \eqref{opt:GenDual}, $ \bm{\Lambda}_d^* $, and confirm that the relaxation gap, $ p(\bm{R}^*) - d(\bm{\Lambda}_d^*) $, is below a small tolerance value. However, even with current state-of-the-art interior-point methods, solving the dual SDP for large problem instances can be prohibitive.

Alternatively, consider the Lagrange multiplier matrix, $ \bm{\Lambda}_p^* $, corresponding to $ \bm{R}^* $. By the Karush-Kuhn-Tucker conditions and the structure of $ \bm{\Lambda}_p^* $, we have
\begin{equation}\label{eqn:LambdaStar}
	(\bm{Q}-\bm{\Lambda}_p^*)\bm{R}^{*T} = \bm{0} 
	\iff \bm{\Lambda}_{p,i}^* = \sum_{j = 1}^{N_p} \bm{Q}_{ij}\mathbf{R}^{*T}_j\mathbf{R}^{*}_i,
\end{equation}
where $ \bm{Q}_{ij} $ refers to the $ ij^{th} $ block of $ \bm{Q} $ when it is partitioned into $ 3\times3 $ matrix blocks. The value of the primal cost function at the candidate optimum is then given by 
\begin{gather*}
	p(\bm{R}^*) = \mbox{tr} \left( \left(\sum_{j = 1}^{N_p}\bm{Q}_{ij}\mathbf{R}^{*T}_j\right)\mathbf{R}^{*} \right)
	= \sum_{i=1}^{N_p} \mbox{tr} \left( \bm{\Lambda}^*_{p,i} \right) = d(\bm{\Lambda}_p^*).
\end{gather*}
We see that the primal and dual objective functions have the same value. Consequently, if it can be shown that $ \bm{\Lambda}_p^* $ is a feasible solution to the dual problem \eqref{opt:GenDual} (i.e., $ \bm{Q}-\bm{\Lambda}_p^* \in \mathbb{S}_+^{3N_p} $), then strong duality must hold. Moreover, the solution $ \bm{R}^* $ must be the global optimum of the primal optimization problem \cite{boyd2004convex}. 

These developments have been demonstrated in other papers \cite{rosen2019se, carlone2015lagrangian, eriksson2018rotation, aholt2012qcqp,chaudhury_global_2015} and a generalization of this result has been presented in Lemma 2.1 of \cite{cifuentes2021local}. It should be mentioned that in all cases we can only find \textit{sufficient} conditions for global optimality so that, in theory, we cannot certify that a candidate local solution is \textit{not} a global solution. However, as shown in Sections \ref{Sec:NumRes} and \ref{Sec:NoiseDuality}, failure of these conditions frequently indicates that a given solution is a local, non-global optimum.

An efficient test for global optimality of $ \bm{R}^* $ is therefore given as follows: once the data matrix, $ \bm{Q} $, has been computed, $ \bm{\Lambda}_p^* $ can be found efficiently in closed form via \eqref{eqn:LambdaStar}. The membership property, $ \bm{Q}-\bm{\Lambda}_p^* \in \mathbb{S}_+^{3N_p} $, can then be determined by computing the minimum eigenvalue of $ \bm{Q}-\bm{\Lambda}_p^* $. The \textit{spectrum shifting} approach presented in \cite{rosen2017computational} leverages Lanczos algorithm \cite{golub2013matrix} and can be used to efficiently find this eigenvalue. Computation of $ \bm{Q} $ therefore constitutes the main computational bottleneck and is addressed in the next section.

\subsection{Efficient Data Matrix Computation}\label{Sec:DataMatComp}
In practical SLAM problems, it is frequently the case that the number of landmarks is large, even in problems with a moderate number of poses. 
In this section, we present a method to compute the data matrix, $ \bm{Q}_{bt}$, with a complexity that remains linear in terms of the number of landmarks\footnote{Note that $ \bm{Q}_r $ is not of concern here since it does not involve landmark measurements in any way.}. Although we show the derivation for a SLAM problem, the method is equally viable for point cloud alignment problems.

Here, we assume an \textit{edge ordering} on $ \EdgeSetB \subset \EdgeSetS  $ such that edges that are adjacent in terms of landmark vertices are grouped together. The incidence matrix, $ \bm{V}_s $, can be partitioned as follows:
\begin{equation}\label{eqn:VsPartStruct}
	\bm{V}_s =
	\begin{bmatrix}
		\bm{V}_p \\
		\bm{V}_m
	\end{bmatrix}=
	\begin{bmatrix}
		\bm{V}_{p,1} & \cdots & \bm{V}_{p,N_m} & \bm{V}_{t} \\
		\bm{V}_{m,1} & \cdots & \bm{V}_{m,N_m} &  \bm{0}
	\end{bmatrix},
\end{equation}
such that $ \bm{V}_{m,i}^T\bm{V}_{m,j} =\bm{0} $ for all $ i\ne j $. This follows from the fact that the subgraph $ \GraphB \subset \GraphS $, corresponding to the restriction of the edge set to  $ \EdgeSetB\subset\EdgeSetS $, is bipartite. 
The projection matrix, $ \bm{A}_s $, appearing in the expression for $ \bm{Q}_{bt} $ can be written in terms of the weighted Laplacian,  $ \bm{L}_s $:
\begin{equation*}
	\bm{A}_s = \bm{I} - \bm{V}^+_s \bm{V}_s = \bm{I} - \bm{V}_s^T\bm{L}_s^+\bm{V}_s,
\end{equation*}
where $ \bm{L}_s $ takes the following form:
\begin{equation}
	\bm{L}_s = \begin{bmatrix}
		\bm{V}_p\bm{V}_p^T & \bm{V}_p\bm{V}_m^T \\
		\bm{V}_m\bm{V}_p^T & \bm{V}_m\bm{V}_m^T
	\end{bmatrix}.
\end{equation}
We note that the \textit{landmark degree} matrix $ \bm{V}_m\bm{V}_m^T $ (denoted hereafter by $ \bm{D} $) is a diagonal matrix due to the partitioning in \eqref{eqn:VsPartStruct} and the fact that the matrix $ \bm{W}_b $ is diagonal\footnote{Alternatively, sparsity may be exploitable in terms of the $ \bm{V}_p\bm{V}_p^T $ submatrix by using a different edge ordering}.

The pseudoinverse, $ \bm{L}_s^+ $, can present difficulties when factorizing $ \bm{Q}_{bt}$ for efficient computations. However, it was shown in Appendix B.3 of \cite{rosen2019se} that the projection matrix, $ \bm{A}_s $, can be replaced by a \emph{reduced} projection matrix,  $\bm{A}'_s = \bm{I} - \bm{V}'^T_s\bm{L}'^{-1}_s\bm{V}'_s $, where $ \bm{V}'_s $ is known as the \emph{reduced} weighted incidence matrix and is obtained by removing the first row of $ \bm{V}_s $. The \emph{reduced} Laplacian, $ \bm{L}'_s=\bm{V}'_s\bm{V}'^T_s $, is known to be full rank and is therefore invertible\footnote{We recapitulate the development presented in \cite{rosen2019se} in Appendix \ref{App:reduced}}. The inverse of $ \bm{L}'_s $ can now be expressed via the Schur complement LDU decomposition,
\begin{multline*}
	\bm{L}'^{-1}_s = \begin{bmatrix}
		\bm{I} & \bm{0} \\
		-\bm{D}^{-1}\bm{V}_m\bm{V}'^T_p & \bm{I}
	\end{bmatrix}\begin{bmatrix}
	(\bm{V}'_p \bm{E} \bm{V}'^T_p)^{-1} & \bm{0} \\
	\bm{0} & \bm{D}^{-1}
\end{bmatrix} \\ 
\begin{bmatrix}
	\bm{I} & -\bm{V}'_p\bm{V}_m^T \bm{D}^{-1}\\
	\bm{0} & \bm{I}
\end{bmatrix},
\end{multline*}
where $ \bm{E} = \bm{I} - \bm{V}_m^T\bm{D}^{-1}\bm{V}_m $ and $ \bm{V}'_p $ is obtained by removing the first row of $ \bm{V}_p $. It follows that
\begin{align*}
	\bm{A}'_s &= \bm{I} - \begin{bmatrix}
		\bm{E}^T\bm{V}'^T_p & \bm{V}_m^T
	\end{bmatrix}
	\begin{bmatrix}
		(\bm{V}'_p \bm{E} \bm{V}'^T_p)^{-1} & \bm{0} \\
		\bm{0} & \bm{D}^{-1}
	\end{bmatrix}
	\begin{bmatrix}
		\bm{V}'_p\bm{E} \\ \bm{V}_m
	\end{bmatrix}\\
	&= \bm{E} - \bm{E}\bm{V}'^T_p(\bm{V}'_p \bm{E} \bm{V}'^T_p)^{-1}\bm{V}'_p\bm{E}.
\end{align*}

\noindent The full data matrix can be written as
\vspace*{-5pt}
\begin{multline}\label{eqn:QbtBlocks1}
		\bm{Q}_{bt} = \underbrace{\bar{\bm{V}}^p_{s} \bm{Y}_s \bm{E} \bm{Y}_s^T\bar{\bm{V}}^{pT}_{s}}_{3N_p\times3N_p}\\
		 - (\underbrace{\bar{\bm{V}}^p_{s} \bm{Y}_s \bm{E}\bm{V}'^T_p}_{3N_p\times N_p}) (\underbrace{\bm{V}'_p \bm{E} \bm{V}'^T_p}_{N_p\times N_p})^{-1}  (\underbrace{\bm{V}'_p\bm{E}\bm{Y}_s^T\bar{\bm{V}}^{pT}_{s}}_{N_p\times 3N_p}).
\end{multline}

We see that the dimensions of the matrices above do not depend on the number of landmarks, $ N_m $. It remains to show that these matrices can be constructed with complexity that is linear in $ N_m $.
Now, owing to the column structure of $ \bm{V}_m $ shown in \eqref{eqn:VsPartStruct}, we have that $ \bm{E} = \mbox{diag}(\bm{E}_1,\dots,\bm{E}_{N_m},\bm{I}) $, where $ \bm{I} $ is the identity matrix of size $ \card{\EdgeSetP} $. The matrix $ \bm{Y}_s $ is block-diagonal by definition, but we partition its blocks similarly, $ \bm{Y}_s = \mbox{diag}(\bm{Y}_1,\dots,\bm{Y}_{N_m},\bm{Y}_{t})$.
We can rewrite the blocks of \eqref{eqn:QbtBlocks1},
\begin{subequations} \label{eqn:QbtBlk}
	\begin{equation}\label{eqn:QbtBlk1}
		\scalebox{0.91}{$\bar{\bm{V}}^p_{s} \bm{Y}_s \bm{E} \bm{Y}_s^T\bar{\bm{V}}^{pT}_{s} = \bar{\bm{V}}^p_{t} \bm{Y}_t \bm{Y}_t^T\bar{\bm{V}}^{pT}_{t} + \sum\limits_{i=1}^{N_m} \bar{\bm{V}}_{p,i} \bm{Y}_{i} \bm{E}_i \bm{Y}_i^T\bar{\bm{V}}^{T}_{p,i},$}
	\end{equation}\vspace*{-3pt}
	\begin{equation}\label{eqn:QbtBlk2}	
		\scalebox{0.91}{$\bar{\bm{V}}^p_{s} \bm{Y}_s \bm{E}\bm{V}'^T_p = \bar{\bm{V}}^p_{t} \bm{Y}_t\bm{V}'^T_t - \sum\limits_{i=1}^{N_m} \bar{\bm{V}}_{p,i} \bm{Y}_i \bm{E}_i\bm{V}'^T_{p,i},$}
	\end{equation}\vspace*{-3pt}
	\begin{equation}\label{eqn:QbtBlk3}
		\scalebox{0.91}{$\bm{V}'_p \bm{E} \bm{V}'^T_p = \bm{V}'_t\bm{V}'^T_t + \sum_{i=1}^{N_m} \bm{V}'_{p,i} \bm{E}_i \bm{V}'^T_{p,i},$}
	\end{equation}
\end{subequations}
where index $ i $ enumerates the blocks in the partition given in \eqref{eqn:VsPartStruct} and $ \bm{V}'_t $ is equivalent to $ \bm{V}_t $ with the first row removed. We have also used the fact that $ \bm{V}^p_{s,i} = -\bm{V}_{p,i}  $. We note that all of the matrix multiplications in \eqref{eqn:QbtBlk} have constant complexity with respect to the number of landmarks, $ N_m $. Therefore, we have shown that the construction of data matrix, $ \bm{Q}_{bt} $, has complexity that is linear in the number of landmarks, $ N_m $. 
All three matrices in this expression are sparse in typical SLAM applications, but explicit computation of the inverse in \eqref{eqn:QbtBlocks1} would make the resulting matrix dense. However, we can efficiently factor the inverted matrix into sparse, lower-triangular matrices, via, for example, a sparse Cholesky decomposition: $(\bm{V}'_p \bm{E} \bm{V}'^T_p)^{-1} = \bm{C}^{-1}\bm{C}^{-T} $. Since our development in this paper only involves \emph{products} with the matrix $ \bm{Q}_{bt} $, we can efficiently store the sparse triangular factor, $ \bm{C} $, and compute the inverse products quickly via substitution.

This construction can be shown to have (worst case) $ O(N_p^3 + N_p^2 N_m) $ complexity. Since $ \bm{Q}-\bm{\Lambda}_p^* $ is a $ 3N_p \times 3N_p $ matrix, its eigenvalues can be computed with (worst case) $ O(N_p^3) $ complexity\footnote{In practice, computing the minimum eigenvalue is much faster than $ O(N_p^3) $ when the Lanczos iteration or other methods are used}, global optimality can be demonstrated with worst case $ O(N_p^3 + N_p^2 N_m) $ complexity, which is the same as that of an iteration of a sparse local solver for Problem \eqref{opt:SLAM} that exploits $ N_m \gg N_p $ \cite{barfoot2017state}\cite{brown1958solution}. Linear complexity with respect to number of landmarks is demonstrated numerically in Section \ref{Sec:SeSyncCompl}.

\section{Numerical Results for Optimality Test}\label{Sec:NumRes}


In this section, we show that the global optimality test described in \ref{Sec:LagrDualCert} can be used to certify landmark-based SLAM problems. Results were applied to both simulated and real-world datasets and all numerical results were generated using Matlab 2021a. A sparse Gauss-Newton (GN) solver was developed in Matlab and used to search for first-order critical points of the landmark-based SLAM problems. The solver was assumed to be \emph{converged} when the square of the step size dropped below $ 1\times10^{-10} $. To compare solutions across trials, one pose was locked to its corresponding ground truth pose. 
For each of the simulated and real-world trials, the data matrix, $ \bm{Q} $, was generated and eigenvalues of $ \bm{Q} - \bm{\Lambda}^*_p $ were checked to verify optimality via the \textit{spectrum shifting} method. To check correctness, the dual SDP \eqref{opt:GenDual} was also set up and solved using CVX with the MOSEK solver \cite{cvx}\cite{mosek}. 

In what follows, we define the \textit{optimality gap} of a primal solution as the difference between primal and dual optimal values, relative to the dual optimal value (i.e., $ (p(\bm{R}^*) - d(\bm{\Lambda}^*_d))/d(\bm{\Lambda}^*_d) $). The optimality gap vanishes for solutions that are globally optimal, while the gap remains large for local minima. Recall that our global optimality test constructs $ \bm{Q} - \bm{\Lambda}_p^* $ via \eqref{eqn:LambdaStar} and computes the minimum eigenvalue via the \textit{spectrum shifting} Lanczos iteration, as described above. A solution was considered to be a global minimum when the $ \bm{Q}-\bm{\Lambda}_p^* $ matrix had a minimum eigenvalue greater than $ -1\times10^{-8} $ (i.e., considered machine zero).

\subsection{Simulated Dataset}\label{Sec:SimData}

A collection of 100 simulated problems were generated with 30 poses and 200 landmarks distributed about an elliptical trajectory (major axis: 15 [m], minor axis: 10 [m]). A pose-graph was generated assuming each pose was connected to the next pose along the trajectory, with the final pose connecting to the first (see Figure \ref{fig:threeMins}). Pose-to-landmark measurements were generated between each pose and all landmarks within 4.5 [m] of that pose. The translation measurements were corrupted by isotropic Gaussian noise (standard deviation: 5 [cm]). The rotational measurements were corrupted by isotropic Gaussian noise in the Lie algebra, $ \mathfrak{so}(3) $ (standard deviation: 10 [deg]). See \cite{barfoot2014associating} for details on this method. 

For each of the 100 problem instances, 10 trials were run with different initial conditions. Of the resulting 1000 trials, 426 solutions converged to global minima. It was observed that the simulated problems had several different local minima, with the shape of each minimum depending on the initial conditions of the GN solver. Please see Appendix \ref{App:LocalMin} for examples of global and local minima found for the simulated dataset.

A solution to a given trial was considered to be a true global minimum if its optimality gap was verified to be near zero (by solving the dual problem). Based on our results, the optimality test was able to \textit{exactly} differentiate between local and global minima of landmark-SLAM with unity precision and recall (defined as in \cite{carlone2015lagrangian}). 

\subsection{Real World Dataset}
The optimality test outlined in Section \ref{Sec:OptCert} was also tested on a segment of the ``Starry Night'' dataset \cite{mactavish2015all}\cite{barfoot2011state} with 715 poses and 20 landmarks. Note that the original measurements in this dataset were in stereo image format and were converted to 3D points as a preprocessing step. The local solver described above was applied to the real-world dataset to obtain one local and one global minimum, based on different initial conditions. These two solutions are shown in Figure \ref{fig:starry} and the certificate results are given in the table below. 
\begin{table}[H]
\vspace*{-3mm}
\centering
\label{Tab:StarryCert}
\resizebox{\columnwidth}{!}{
\begin{tabular}{|c|c|c|c|c|}
	\hline
	& \textbf{Cost} & \textbf{Optimality} & \textbf{Min. Eig. of} & \textbf{Cert.}  \\
	&&\textbf{Gap}&$ \bm{Q}-\bm{\Lambda}_p^* $& \\
	\hline
	\textbf{Global Min.} & $ 7.4\E{-2} $ &$ -6.3\E{-5} $&$ -2.0\E{-14} $& PASS\\ 	\hline
	\textbf{Local Min.} &1.88 & 24.5 & $ -5.7\E{-3} $ & FAIL \\
	\hline
\end{tabular}}
\vspace*{-3mm}
\end{table}
From the table, it is clear that the certificate successfully differentiates global minima from local minima\footnote{For this dataset, the dual SDP was also solved using CVX with MOSEK with optimal value $ 7.4\E{-2} $ and was used to define optimality gap.}.
%
%
%

\section{Integration with SE-Sync and Complexity Results}\label{Sec:SeSyncCompl}

As mentioned above, it is possible to use SE-Sync to solve and certify landmark-based SLAM problems by treating landmark variables as ``degenerate'' poses. That is, the problem can be solved as if each landmark is a pose with both translation and (dummy) rotation variables. However, depending on the number of landmarks (which can be large in typical SLAM problems), this naive approach can prohibitively increase the size of the problem that must be solved. 

We present an alternative approach, wherein we first compute an implicit representation of the matrix $ \bm{Q}_{bt} $ via \eqref{eqn:QbtBlocks1} and combine it with $ \bm{Q}_r $ in \eqref{eqn:Qr}. We then integrate this implicit representation into the SE-Sync pipeline to compute the certified solution $ \bm{R}^* $ and recover the pose translations and landmarks via \eqref{eqn:transSoln}. This, in effect, marginalizes the landmark information to keep the problem size small while still maintaining the original sparsity of the problem\footnote{Our implementation of this approach is available at \url{https://github.com/holmesco/SE-Sync-Landmarks}.}.

The naive SE-Sync approach to the landmark-based SLAM problem was compared to our marginalized approach in terms of overall runtime and runtime of particular components of SE-Sync. This comparison was performed in MATLAB 2019a using the simulated data outlined in Section \ref{Sec:SimData}, but with the number of poses set to 100 and the number of landmarks left as a variable in the comparison. The parameters for the SE-Sync algorithm were set based on the provided minimum working example \cite{rosen2019se}.

\begin{figure}[!t]
	\centering
	\vspace*{-10pt}
	\includegraphics[width=\columnwidth]{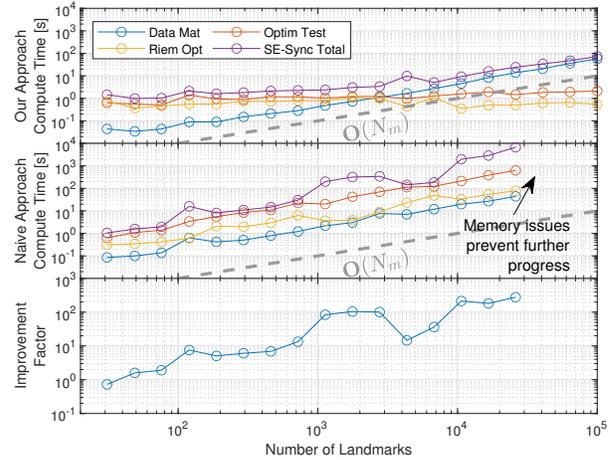}
	\vspace*{-25pt}
	\caption{Comparison of SE-Sync runtime when landmark variables are introduced. First plot show our marginalized approach into SE-Sync while the second plot shows the naive approach, representing each landmark by a ``degenerate'' pose. The final plot shows the ratio of naive runtime to our runtime: an \textbf{order of magnitude} improvement is seen for even modest numbers of landmarks (e.g., $ N_m = 100 $). Data matrix generation (blue) is $ \mbox{O}(N_m) $ for both approaches. Riemannian optimization (yellow) and certification (red) have nearly constant runtime in our approach, but are $ \mbox{O}(N_m) $ in the naive approach. The naive approach could not run up to the maximum number of landmarks due to a memory issue. }
	\vspace*{-15pt}
	\label{fig:SeSyncComp}
\end{figure}

The resulting comparison is shown in Figure \ref{fig:SeSyncComp}. Interestingly, both our marginalized approach and the naive SE-Sync approach also appears to have complexity that is linear in the number of landmarks. However, our approach offers an \emph{order of magnitude} improvement over the naive approach in terms of total runtime (purple in Figure \ref{fig:SeSyncComp}). Upon closer inspection, the reason for this becomes clear: the Riemannian optimization (yellow) and Lanczos iteration optimality test (red) scale linearly with the landmarks in the naive approach, but have \emph{constant} complexity in our marginalized approach\footnote{Note that in some cases, the total SE-Sync time is much higher than any of the composite steps. It was observed that this occurred specifically when the algorithm ``ascends'' the Riemannian staircase (see \cite{rosen2019se}).}. 

\section{Noise Bounds for Strong Duality}\label{Sec:NoiseDuality}

It has been shown, for both general QCQPs \cite{cifuentes2021local} and related QCQP perception problems \cite{rosen2019se,carlone2015lagrangian,iglesias_global_2020}, that strong duality depends on the level of measurement noise present in a given problem. It is important to understand this dependency because if strong duality can be confirmed or safely assumed, then the global optimality test described in the preceding sections becomes both \emph{sufficient} and \emph{necessary} for global optimality.

In this section, we provide a numerical study of the maximum noise level for which strong duality still holds for the landmark-based SLAM QCQP. We call this level the \emph{noise-duality bound} (NDB) and also study the effects of different aspects of the measurement graph $ \GraphS $ on this bound. In particular, we consider the number of landmarks and the connectivity of $ \GraphS $ in terms of both number of pose-to-pose loop closures and number of pose-to-landmark measurements as a fraction, $ \alpha $, of the fully connected case.

All of the figures in this section are based on randomly generated datasets of 20 poses and varying numbers of landmarks within a bounding box of 0 [m] to 50 [m] on each axis. The baseline noise for the measurements was set to a standard deviation of 0.866 [m] for translation measurements and 0.573 [deg] for rotation measurements. These standard deviations were scaled to determine the noise value at which the duality gap starts to break down. We use the average corank of $ \bm{H}_d $ across 30 trials as an indicator of strong duality. For each trial, the matrix $ \bm{H}_d $ is computed by solving the dual SDP \eqref{opt:GenDual} using MOSEK. 

In all figures, it is clear that the noise levels for which strong duality breaks down in landmark-based SLAM are quite high. While the baseline noise levels are already high, this study indicates that a duality gap is not apparent until at least another factor of 10 above the baseline. This has also been noted in similar robotics and vision QCQPs (e.g., \cite{carlone2015lagrangian,eriksson2018rotation,rosen2019se,iglesias_global_2020}).

\begin{figure}[!t]
	\centering
	\vspace*{-0pt}
	\includegraphics[width=\columnwidth]{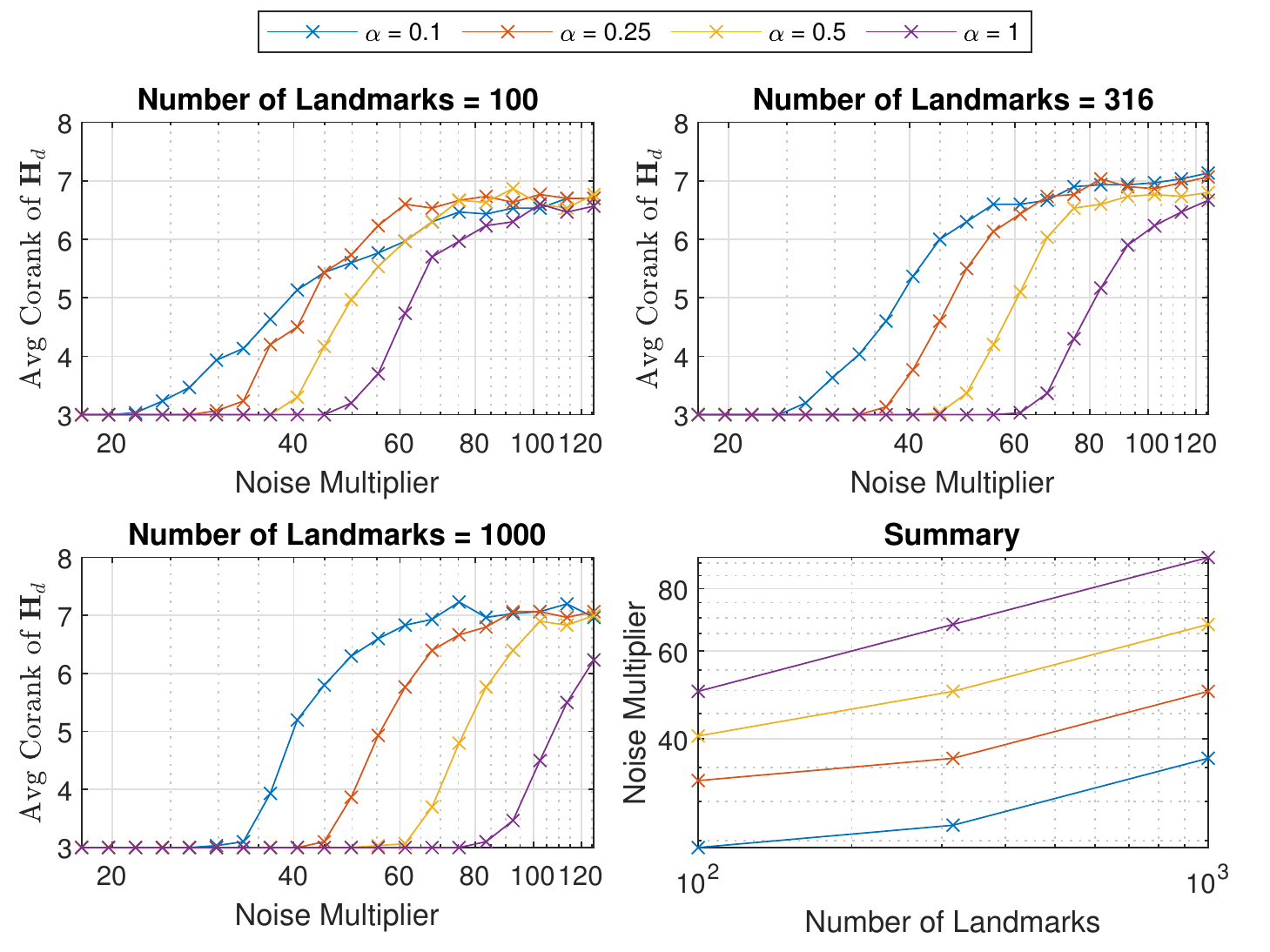}
	\vspace*{-20pt}
	\caption{Effect of the number of landmarks on the NDB when the pose-to-pose measurement edges remain fixed. $ \alpha $ refers to the fraction of pose-to-landmark edges allowed in the measurement graph with respect to a fully connected graph. In other words, $ \alpha $ represents the connectivity of the pose-to-landmark part of $ \GraphS $. The summary plot shows the point at which the average corank of $ \bm{H}_d $ exceeds 3 (meaning strong duality does not hold for at least one trial).}
	\vspace*{-15pt}
	\label{fig:LmGraphStudy}
\end{figure}

\begin{figure}[!b]
	\centering
	\vspace*{-20pt}
	\includegraphics[width=\columnwidth]{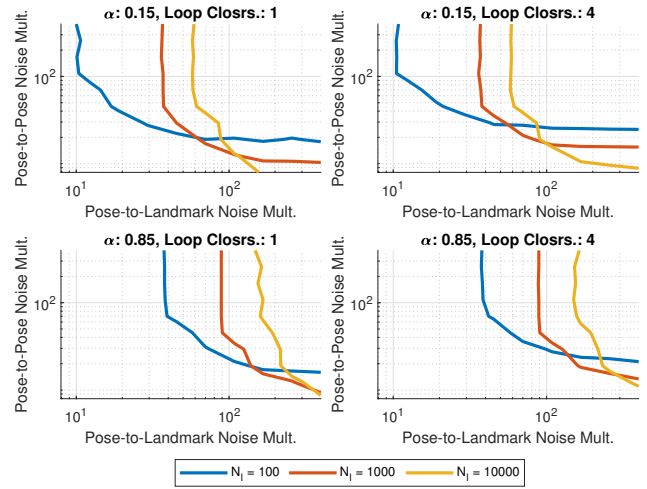}
	\vspace*{-20pt}
	\caption{Study of the effects pose-to-pose and pose-to-landmark noise while also considering number of landmarks ($ N_l $), connectivity of the pose-to-landmark graph ($ \alpha $), and number of loop closures. Contours show the boundary at which noise levels cause strong duality to break down (corank of $ \bm{H}_d > 3$) for at least half of the trials. Note that strong duality holds for regions on the low noise side of the boundary.}
	\vspace*{-0pt}
	\label{fig:SepNoiseStudy}
\end{figure}

It was shown in \cite{iglesias_global_2020} that removing pose-to-landmark edges for a fixed number of landmarks reduces the NDB. Figure \ref{fig:LmGraphStudy} reproduces this result when pose-to-pose measurements are also included (i.e., for landmark-based SLAM) and shows the effect of varying the overall number of landmarks. This figure demonstrates that increasing the number of landmarks and increasing connectivity of the pose-to-landmark graph ($ \alpha $) cause the NDB to increase when all noise levels are scaled simultaneously. This is not true in general, as Figure \ref{fig:SepNoiseStudy} demonstrates (to be discussed shortly).

\begin{figure}[!t]
	\centering
	\vspace*{-0pt}
	\includegraphics[width=\columnwidth]{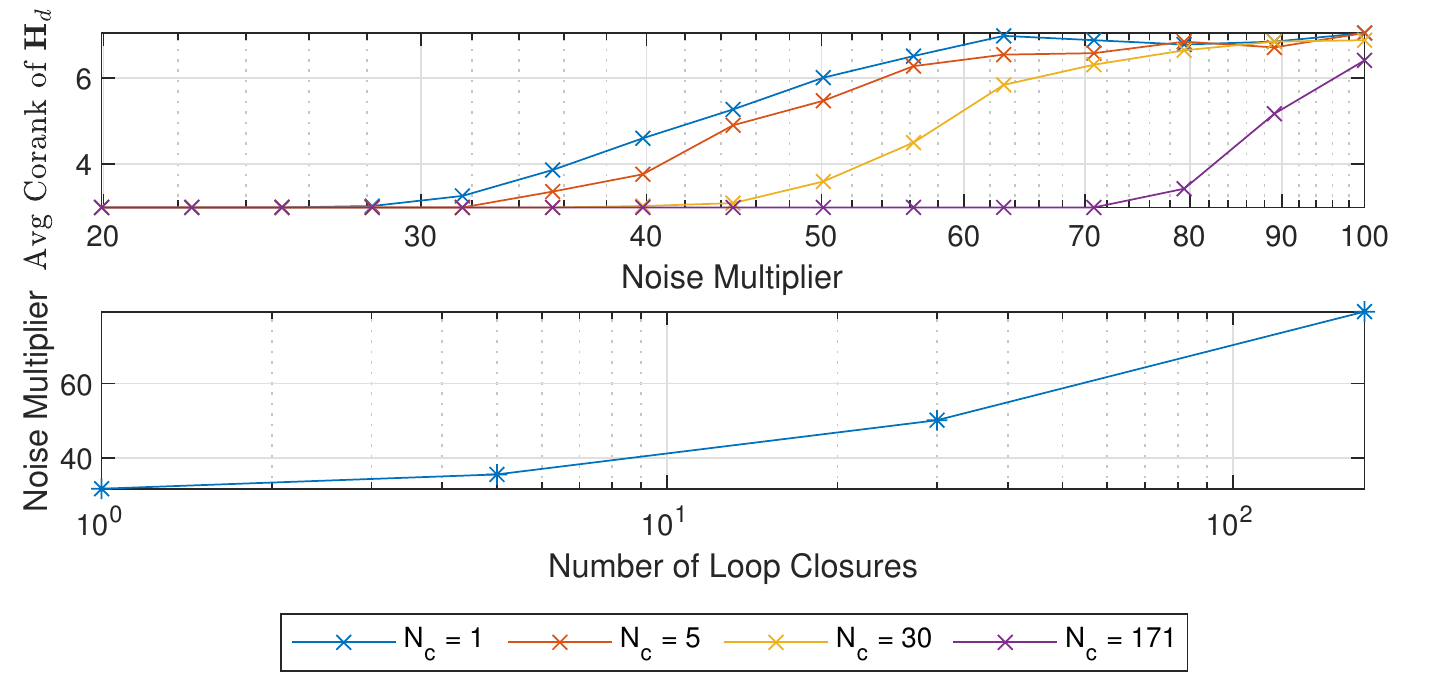}
	\vspace*{-20pt}
	\caption{Effect of loop closures in the pose-to-pose measurement graph with $ 200 $ landmarks and pose-to-landmark fraction $ \alpha = 0.2 $. Upper plot shows the average corank of $ \bm{H}_d $ for different numbers of loop closures, $ N_c $. Lower plot shows the noise level for which the average corank exceeds 3 (at least one case of strong duality failure). For the number of poses considered, 171 corresponds to the maximum number of distinct loop closures.}
	\vspace*{-20pt}
	\label{fig:LoopCloseStudy}
\end{figure}

In Figure \ref{fig:LoopCloseStudy}, we consider the effect of pose-to-pose measurements on the NDB when the pose-to-landmark measurements are fixed. Specifically, consecutive poses are linked by measurements (as they would be for a SLAM instance) and we show the changes to the NDB as `loop closure' measurements are added. As one might expect, the addition of loop closures improves the NDB.

Finally, in Figure \ref{fig:SepNoiseStudy}, we study the noise level of pose-to-pose and pose-to-landmark measurements separately, while also considering other factors. Interestingly, increasing the number of landmarks improves the bound when pose-to-landmark noise is low, but has the opposite effect when pose-to-landmark noise is high. A similar effect can be seen for the connectivity of the pose-to-landmark graph. 

We interpret this effect as follows: as more landmarks are added, the data matrix $ \bm{Q} $ for the landmark-based SLAM problem becomes more `dominated' by the landmark measurements. Therefore, when the number of landmarks is high, the pose-to-landmark noise level has a large influence over the NDB. Similarly, when the connectivity of the pose-to-landmark graph increases, the data matrix is, again, more dominated by the pose-to-landmark measurements.

Since the number of poses considered here is relatively small (20 poses), we only considered up to 4 loop closures. These closures seem to have very little effect on the bound when the pose-to-pose noise level is high and improve the bound when pose-to-landmark noise levels are high. These observations are consistent with our interpretation.

In future work, it is our intention to use the trends shown in this section to guide a theoretical derivation of measurement noise bounds for which strong duality can be guaranteed, following similar analyses in \cite{eriksson2018rotation}, \cite{rosen2019se} and \cite{cifuentes2021local}.

\section{Conclusion}
In this paper, we have shown that the landmark-based SLAM problem --- along with its subproblems --- can be expressed in a common form that admits \textit{sufficient} conditions for global optimality. We have shown the \textit{data matrix} can be computed and a global optimality certificate obtained with complexity that is linear in the number of landmarks with worst case complexity equal to that of a single local solver iteration. We have also verified that, for simulated and real-world landmark-based SLAM problems, the optimality conditions outlined in this paper consistently and correctly identify global minima. We integrated our method into the existing SE-Sync pipeline and showed that this integration improves runtime by an order of magnitude even when the number of landmarks is moderate. 
Finally, we have investigated how the level of noise and aspects of the measurement graph affect the strong duality of the problem. In the future, we hope to use the results shown in Section \ref{Sec:NoiseDuality} to guide a theoretical analysis of the NDB, leveraging theorems from \cite{cifuentes2021local,rosen2019se}. Other future directions for this work involve including matrix weight factors in the cost function and accounting for outlier measurements as in \cite{yang2020teaser}.
\section{Acknowledgments}
The authors would like to thank Luca Carlone for helpful discussions in the preparation of this paper and Frederike D{\"u}mbgen for her insightful comments. 

\bibliography{CertifiablePerception,CertificationRefs,SemidefiniteProgramming}
\bibliographystyle{IEEEtrans}

\renewcommand\appendixname{Appendix (arXiv Only)}
\appendix
\numberwithin{equation}{section}
\renewcommand{\theequation}{\thesection.\arabic{equation}}

\subsection{Lagrangian Duality and SDP Relaxation} \label{App:LagrDual}
In this section, we derive the Lagrangian dual and SDP relaxation of Problem \eqref{opt:SLAM}. For each constraint ($ \bm{R}_{oi}\bm{R}_{oi}^T = \bm{I} $) in Problem \eqref{opt:GenSDP}, we define the symmetric Lagrange multiplier, $ \bm{\Lambda}_i \in \mathbb{S}^3 $. For convenience, these multipliers can be collected into the matrix $ \bm{\Lambda} = \mbox{diag}(\bm{\Lambda}_1,\dots,\bm{\Lambda}_{N_p})$. The Lagrangian function for Problem \eqref{opt:GenSDP} is defined as follows:
\begin{align}
	\mathcal{L}(\bm{R},\bm{\Lambda}) &=  \mbox{tr} \left(\bm{Q} \bm{R}^T \bm{R} + \bm{\Lambda}^T (\bm{I}-\bm{R}\bm{R}^T) \right) \nonumber\\
	&=  \mbox{tr} \left( \bm{R}(\bm{Q}-\bm{\Lambda}) \bm{R}^T + \bm{\Lambda}\right).
\end{align}
The Lagrangian dual function, $ d(\bm{\Lambda}) $, is therefore given by the following (unconstrained) minimization:
\begin{align}
	d(\bm{\Lambda}) &= \inf_{\bm{R}} \quad \mbox{tr} \left( \bm{R}(\bm{Q}-\bm{\Lambda}) \bm{R}^T + \bm{\Lambda}\right) \nonumber \\
	&= \left\{ \begin{array}{ll}
		\mbox{tr}(\bm{\Lambda}) \quad& \mbox{if } (\bm{Q}-\bm{\Lambda}) \succeq \bm{0} \\
		-\infty & \mbox{otherwise}
	\end{array}\right.,
\end{align}
where $ \succeq $ denotes inequality with respect to the positive semidefinite cone. The dual function is only finite when $ (\bm{Q}-\bm{\Lambda}) \succeq \bm{0} $. Therefore, the Lagrangian dual problem can be written as
\begin{equation}\label{opt:GenDual2}
	\max\limits_{\bm{\Lambda}} d(\bm{\Lambda}) =\mbox{tr}(\bm{\Lambda}), \quad \mbox{s.t.}~ \bm{H} \coloneqq(\bm{Q}-\bm{\Lambda}) \succeq \bm{0}.
\end{equation}
We note that this problem is a SDP (convex) and the feasible set of this program has non-empty interior for any $ \bm{Q} $ (i.e., Slater's condition is satisfied). Therefore, Problem \eqref{opt:GenDual2} admits a global maximum, $ \bm{\Lambda}^*_d $ \cite{boyd2004convex}. Moreover, strong duality for the original QCQP \eqref{opt:GenSDP} holds whenever the corank of the matrix $ \bm{H}_d \coloneqq \bm{Q} - \bm{\Lambda}^*_d $ is 3.
To see why this is true, consider the dual of the dual problem \eqref{opt:GenDual2}, which is equivalent to the rank relaxation SDP of \eqref{opt:GenSDP}:
\begin{equation}\label{opt:GenSDPRelax2}
	\begin{array}{rl}
		\min\limits_{\bm{Z}} & \mbox{tr} \left(\bm{Q} \bm{Z}\right)\\
		\mbox{s.t.} & \left[\bm{Z}_{ii}\right] = \bm{I}, \quad (\forall i = 1,\dots,N_p),\\
		& \bm{Z} \succeq \bm{0},
	\end{array}
\end{equation}
where $ \bm{Z}_{ii} $ refers to the $ i^{th} $ diagonal block of $ \bm{Z} $ when it is partitioned into $ 3\times3 $ matrix blocks. By the Karush-Kuhn-Tucker conditions, $ \bm{H}\bm{Z}^*=\bm{0} $ for optimal solution $ \bm{Z}^* $ and therefore if the corank of $ \bm{H} $ is 3, the rank of $ \bm{Z}^* $ is at most 3\footnote{$ \mbox{rank}(\bm{Z}) + \mbox{rank}(\bm{H}) = 3N_p $ is equivalent to the strict complementarity condition, which holds generically for SDPs.} \cite{alizadeh_complementarity_1997}.
Strong duality holds for \eqref{opt:GenSDPRelax2} and, if the rank of $ \bm{Z}^* $ is 3, we can decompose as $ \bm{Z}^* = \bm{R}_Z^*\bm{R}_Z^{*T}$, where $ \bm{R}_Z^* $ is now a solution to \eqref{opt:SLAM} for which strong duality holds. 

\subsection{Shape of Local Minima in SLAM Problems}\label{App:LocalMin}

\begin{figure}[H]
	\centering
	\vspace*{-0in}
	\includegraphics[width=\columnwidth]{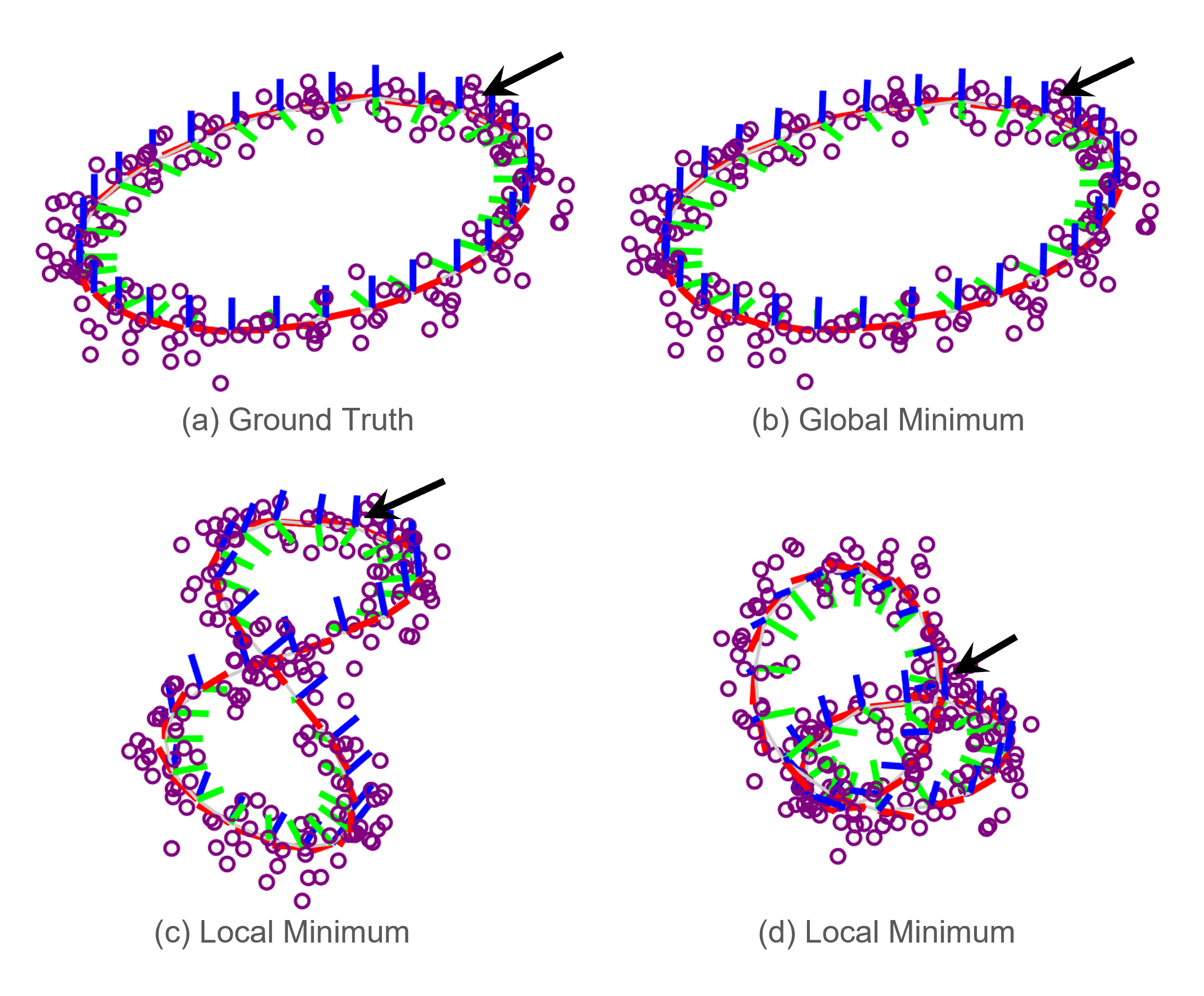}
	\vspace*{-20pt}
	\caption{Ground truth, (a), and three local minima, (b)-(d), for a landmark-based SLAM problem. Note that the local minima were obtained via Gauss-Newton iterations on the same measurement set, but with different initial conditions. The optimal dual value, $ d(\bm{\Lambda}_d^*)$, was 1531.8 and primal values, $ p( \bm{R}^*) $, for each solution were: (b) 1531.8, (c) 2094.4, (d) 1998.5. Locked pose is indicated by black arrows.}
	\vspace*{-5pt}
	\label{fig:threeMins}
\end{figure}

Figure \ref{fig:threeMins} shows one problem instance from Section \ref{Sec:SimData} in which three distinct minima were observed. It is interesting to note that the twisted local minima trajectories seem to be topologically equivalent to the global minimum trajectory, with small local deformations leading to a large change in shape. This idea resonates with the network-of-springs analogy of SLAM \cite{durrantwhyte06a}, if one considers the minima to be stable shapes into which a flexible structure (a ring in this case) can be deformed. The global minimum corresponds to the shape with the lowest potential spring energy.

\subsection{Reduced Incidence Matrix and Laplacian}\label{App:reduced}

In this section, we show that the projection matrix, $ \bm{A}_s $, and the \textit{reduced} projection matrix, $ \bm{A}'_s $, presented in Section \ref{Sec:DataMatComp} represent the same mapping and can therefore be interchanged, without loss of generality. The development presented herein follows closely the derivation provided in Appendix B.3 of \cite{rosen2019se}.

We first recall that $\bm{A}_s = \bm{I} - \bm{V}^+_s \bm{V}_s $ is a matrix representing an orthogonal projection operator, $ \pi(v): \mathbb{R}^{\card{\EdgeSetS}} \mapsto \ker(\bm{V}_s) $, onto the kernel of the weighted incidence matrix, $ \bm{V}_s $. Now, by the Fundamental Theorem of Linear Algebra, we have
\begin{equation*}
	\ker(\bm{V}_s)^{\perp} = \mbox{image}(\bm{V}^T_s).
\end{equation*}
Since the incidence matrix is associated with the \emph{weakly-connected} graph $ \GraphS $, it is well known that the rank of $ \bm{V}_s $ is $\card{\VertSetS}-1 = N_p + N_m - 1 $. Let $ \bm{V}'_s $ be the weighted incidence matrix, but with the first row removed. Then $ \bm{V}'_s $ has full row rank. Moreover, $  \mbox{image}(\bm{V}^T_s) =  \mbox{image}(\bm{V}'^T_s) $ and, since we are dealing with finite vector spaces, it follows that
\begin{equation*}
	\ker(\bm{V}_s)^{\perp} = \ker(\bm{V}'_s)^{\perp} \iff \ker(\bm{V}_s) = \ker(\bm{V}'_s).
\end{equation*}
This implies that we can equivalently represent $ \pi(v) $ with the following projection matrix:
\begin{align*}
	\bm{A}'_s = \bm{I} - \bm{V}'^+_s \bm{V}'_s &= \bm{I} - \bm{V}'^T_s (\bm{V}'_s\bm{V}'^T_s) \bm{V}'_s \\
	&= \bm{I} - \bm{V}'^T_s (\bm{L}'_s)^{-1} \bm{V}'_s,
\end{align*}
where $ \bm{L}'_s = \bm{V}'_s\bm{V}'^T_s $ is full rank and invertible since $ \bm{V}'_s $ has full row rank.

\subsection{Runtime to Number of Pose}
In this section, we empirically investigate the runtime complexity of the data matrix construction with respect to the number of poses. This comparison was performed in MATLAB 2019a using the simulated data similar to the data outlined in Section \ref{Sec:SimData}, but with the number of landmarks set to 100 and the number of poses left as a variable in the comparison. Using this simulated data, we test our approach to SE-Sync, which includes computation of the data matrix. The parameters for the SE-Sync algorithm were set based on the provided minimum working example in \cite{rosen2019se}.

\begin{figure}[H]
	\centering
	\vspace*{-0.1in}
	\includegraphics[width=\linewidth]{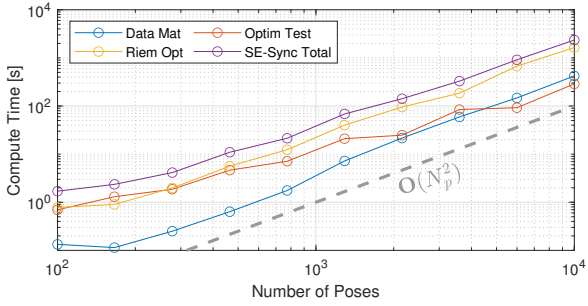}
	\vspace*{-0.15in}
	\caption{This plot demonstrates runtime of our marginalized approach to SE-Sync as the number of poses are increased and the number of landmarks remains fixed. Empirically, it can be seen that the data matrix computation (blue), Riemannian optimization (yellow) and total runtime (purple) have $ \mbox{O}(N_p^2) $ complexity.}
	\label{fig:PoseRuntime}
\end{figure}

Figure \ref{fig:PoseRuntime} confirms the theoretical complexity derived in Section \ref{Sec:DataMatComp}. That is, the time required to compute the data matrix, $ \bm{Q}_{bt} $ is proportional to the square of the number of poses, $ N_p $. Interestingly, we see that the bottleneck step in the modified SE-Sync algorithm is the Riemannian Optimization, which also scales with the square of $ N_p $.

\end{document}